\documentclass[letterpaper]{article} 
\usepackage{aaai2026}  
\usepackage{times}  
\usepackage{helvet}  
\usepackage{courier}  
\usepackage[hyphens]{url}  
\usepackage{graphicx} 
\usepackage{natbib}  
\usepackage{caption} 
\usepackage{algorithm}
\usepackage{algorithmic}
\usepackage{amsmath}
\newtheorem{definition}{Definition}
\newtheorem{theorem}{Theorem}
\newtheorem{example}{Example}
\newtheorem{lemma}{Lemma}
\newtheorem{corollary}{Corollary}
\newcommand{\oprof}[1]{#1_{O}}
\newcommand{\pprof}[1]{#1_{P}}
\usepackage{newfloat}
\usepackage{listings}
\DeclareCaptionStyle{ruled}{labelfont=normalfont,labelsep=colon,strut=off} 
\floatstyle{ruled}
\newfloat{listing}{tb}{lst}{}
\floatname{listing}{Listing}
\title{Beyond Verification: Abductive Explanations for Post-AI Assessment of Privacy Leakage}
\author {
    Belona Sonna\textsuperscript{\rm 1},
    Alban Grastien\textsuperscript{\rm 2},
    Claire Benn\textsuperscript{\rm 3}
}
\affiliations {
    \textsuperscript{\rm 1} Australian National University \\
    \textsuperscript{\rm 2} Université Paris-Saclay, CEA, List\\
    \textsuperscript{\rm 3}Leverhulme Centre for the Future of Intelligence, University of Cambridge\\
    belona.sonna@anu.edu.au, alban.grastien@cea.fr, cmab3@cam.ac.uk
}

\begin{document}

\maketitle

\begin{abstract}
Privacy leakage in AI-based decision processes poses significant risks, particularly when sensitive information can be inferred. We propose a formal framework to audit privacy leakage using abductive explanations, which identifies minimal sufficient evidence justifying model decisions and determines whether sensitive information disclosed. Our framework formalizes both individual and system-level leakage, introducing the notion of \emph{Potentially Applicable Explanations (PAE)} to identify individuals whose outcomes can shield those with sensitive features. This approach provides rigorous privacy guarantees while producing human-understandable explanations, a key requirement for auditing tools. Experimental evaluation on the German Credit Dataset illustrates how the importance of sensitive literal in the model decision process affects privacy leakage. Despite computational challenges and simplifying assumptions, our results demonstrate that abductive reasoning enables interpretable privacy auditing, offering a practical pathway to reconcile transparency, model interpretability, and privacy preserving in AI decision-making.
\end{abstract}


\section{Introduction}
Artificial intelligence (AI) and machine learning (ML) models are increasingly central to critical decision processes in fields such as healthcare, finance, and public administration. Their utility stems from the ability to analyze complex data. At the same time, reliance on opaque ``black-box'' architectures introduces significant challenges, particularly regarding transparency, accountability, and privacy. Specifically, privacy leakage—defined as the unintended exposure of sensitive information through the outputs, behavior, or infrastructure of an AI system~\cite{Sakib2023MeasuresOI}—has become a critical concern. In post-deployment settings, models often interact with continuously evolving data, sometimes in adversarial environments, yet existing practices struggle to systematically assess or audit emerging privacy risks~\cite{RASHEED2022106043}. These risks, combined with heightened regulatory requirements and public concern further underscore the need for verifiable, rigorous assurances that sensitive information handled by these systems is not inadvertently exposed~\cite{Murphy2021CORRSW}.

In response to these demands, the field of explainable AI (XAI) has grown rapidly, offering tools to improve interpretability\footnote{Interpretability refers to the extent to which a human can comprehend the internal mechanics of a machine learning system~\cite{lipton2018mythos}.} and user trust~\cite{A2023100230}. Techniques such as feature attribution, counterfactual explanations, and local surrogate models have become standard for interpreting model behavior. Nonetheless, these approaches often fall short when robust privacy guarantees are required. Feature attributions, for instance, frequently lack minimality and formal correctness, limiting their suitability for audit-ready privacy assurance~\cite{electronics8080832}. Counterfactual explanations, while intuitive, can be highly sensitive to modeling assumptions, providing little guarantee that all sources of privacy leakage are captured or that explanations are unique and complete~\cite{Riccardo}. Similarly, surrogate models trained on local approximations can oversimplify the complexity of high-dimensional domains, such as healthcare or finance, potentially overlooking subtle but critical leakage pathways~\cite{köhler2024achievinginterpretablemachinelearning}.

Given these limitations, abductive explanations have emerged as a compelling alternative. Abductive inference, grounded in formal logic, traces observed outcomes back through the decision process to identify minimal sufficient causes, whether they are input features, model substructures, or adversarial influences~\cite{ijcai2020p726}. By ensuring explanations are both minimal and sufficient, abductive methods support reproducible and auditable reasoning in post-deployment environments. These properties fill a critical gap in privacy accountability, particularly in dynamic settings where empirical or heuristic explanations fail to provide rigorous assurance~\cite{electronics8080832}.

The practical value of abductive reasoning extends beyond theoretical appeal. In bias and fairness auditing, for example, abduction has been used to systematically trace the roots of unfair decisions, enabling formal verification of model behavior across sensitive use cases~\cite{Adnan2020Reasons,sonna2025explainingproxydiscriminationunfairness}. This approach not only highlights weaknesses in heuristic methods but also guides mitigation strategies that directly target the underlying causes revealed by logical inference.

While computational costs of abductive methods in large-scale systems remain a research challenge, advances in symbolic reasoning and optimization are making it increasingly feasible to integrate logic-based explanations with scalable AI deployment. Consequently, a growing toolkit is emerging for sustainable, auditable privacy verification in deployed AI systems, which is crucial for domains where post hoc accountability is not merely desirable but essential~\cite{electronics8080832,ijcai2020p726}.

Overall, sensitive information can be inferred from model outputs, creating substantial risks of privacy leakage. This risk is further amplified by \emph{privacy opacity}, where even system designers or auditors cannot fully predict or understand how sensitive data is processed within the model. Motivated by this challenge, the core question we investigate is: how can abductive reasoning be leveraged to formally detect, explain, and potentially mitigate privacy leakage, ensuring that sensitive information remains protected while maintaining model interpretability and utility?

The remainder of this paper addresses this question as follows: Section~\ref{sec::definition} defines the core concepts of our framework and illustrates how they are applied to analyze privacy leakage, then follow The contributions which are threefold:
\begin{enumerate}
    \item We formally define \emph{individual privacy leakage} and propose a method for auditing it using formal abductive explanations (Section~\ref{sec::individual_level_audit}).
    \item We extend this methodology to the \emph{system level} by assessing privacy leakage across entire decision processes (Section~\ref{sec::global_level_audit}).
    \item Through targeted experimental studies, we empirically investigate how sensitive attributes influence the nature and severity of privacy leakage, offering new insights for privacy risk management (Section~\ref{sec::experimental}).
\end{enumerate}

\section{Background}\label{sec::background}
Privacy opacity\footnote{Privacy opacity refers to the limited visibility outsiders have into personal data or system operations.} is increasingly complex, shaped by digital technologies, ethical imperatives, and societal trust. Traditionally, privacy provided individuals with protective zones against external intrusions, mirrored in legal frameworks balancing personal opacity and institutional transparency~\cite{Gutwirth_DeHert_2022}. However, artificial intelligence and big data introduce new forms of opacity, from users not understanding data use to “deep opacity,” where even experts cannot predict system behavior, undermining informed consent and effective privacy protection~\cite{müller2025deepopacityaithreat}. In sensitive sectors like healthcare, algorithmic opacity restricts patient agency, diminishes trust, and complicates regulatory enforcement~\cite{krista2020}. Although regulatory and technical advances such as explainable AI, homomorphic encryption, and differential privacy have emerged in response, vague privacy disclosures and limited transparency in AI based decision processes highlight persistent gaps~\cite{10628806}.

These challenges underscore the need for systematic auditing of privacy leakage in AI-based decision processes, where maintaining accountability requires balancing two often conflicting goals: transparency\footnote{Transparency in AI refers to the clarity and openness with which AI systems' operations, decisions, and underlying mechanisms are made understandable to stakeholders~\cite{Mitchell2025}.} and data privacy\footnote{Data privacy refers to technical, legal, and organizational measures ensuring that personal, sensitive, or proprietary information processed by AI systems is only accessible to authorized parties and protected from misuse or breaches~\cite{Mohamed2025TheIO}.}. 
In practice, efforts to address this tension have followed two main directions: \emph{ad hoc} methods, which aim to prevent privacy risks during model design, and \emph{post hoc} approaches, which seek to detect and audit leakage after deployment.

Ad-hoc privacy-preserving methods, including federated learning, blockchain-enabled frameworks, and domain-specific machine learning algorithms, reduce privacy risks by avoiding centralized data aggregation and securing communications~\cite{Abdul2022,Iqra2025,Shazia2023}. While effective in specific contexts such as vehicular networks or IoT systems, these methods often incur high computational costs, scalability issues, and reduced interpretability, thereby limiting their applicability in broader settings~\cite{Sanaz2022,Shaik2025}.

Post-hoc explanation techniques address some limitations by providing transparency without altering model architecture~\cite{Fatima2024,Pengrui2022}. Methods like feature importance rankings, counterfactual explanations, and saliency maps allow stakeholders to understand AI decision rationales. However, such techniques can inadvertently expose sensitive dependencies, creating vulnerabilities to membership inference or model extraction attacks. Differential privacy and noise injection partially mitigate these risks but often degrade explanation fidelity\footnote{
Explainability fidelity refers to the degree to which an explanation faithfully represents the true reasoning or decision-making process of the underlying model~\cite{10.1145/2939672.2939778}.} and fail to provide formal, auditable guarantees~\cite{Naoise2019}.

Abductive inference offers a principled alternative by reasoning about the minimal evidence required to justify model decisions~\cite{Vinay2023}. Unlike heuristic or gradient-based methods, abductive frameworks systematically identify structural vulnerabilities and sensitive features, enabling formal guarantees regarding privacy leakage. By combining interpretability with rigorous privacy analysis, abductive reasoning bridges the gap left by conventional approaches, reconciling transparency with confidentiality and supporting accountable AI deployment.
\section{Preliminaries and Description of the Framework}\label{sec::definition}
\subsection{What Information could be safely disclosed: Data Privacy Dilemma}
\begin{figure}[!b]
\centering
\includegraphics[width=0.5\textwidth]{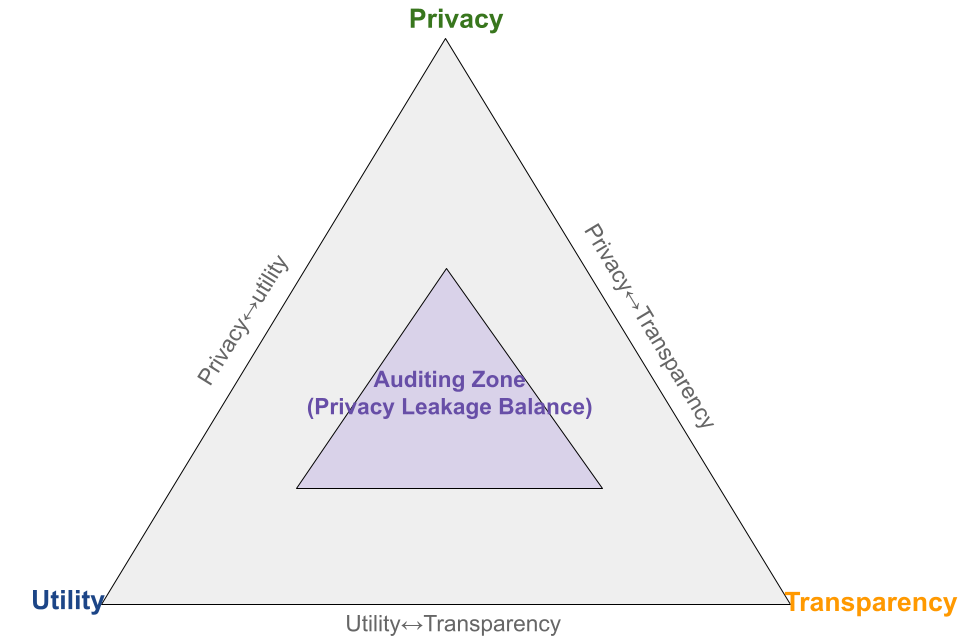}
\caption{The Data Privacy Dilemma: balancing utility, transparency, and privacy in AI-based decision processes}
\label{fig:dillema}
\end{figure}
The figure~\ref{fig:dillema} illustrates the inherent tension between Utility, Transparency, and Privacy in AI-based decision-making systems. Achieving high utility typically requires access to detailed personal data, but such data increase the risk of privacy leakage, which is the unintended exposure of sensitive or identifiable information~\cite{Goryawala2024}. At the same time, ensuring transparency and explainability is essential for accountability, yet it can inadvertently reveal private aspects of the decision rationale~\cite{Zhang2022}. These competing objectives form a triangular trade-off, often referred to as the \emph{Data Privacy Dilemma}~\cite{s23031151}, in which optimizing one dimension can undermine another.

At the center of this triangle lies the \emph{Auditing Zone}, representing the safe zone where privacy auditing tools aim to operate: sensitive information is kept confidential, while sufficient information is disclosed to preserve model utility and interpretability. The primary objective of these tools is to systematically identify which elements of the decision process can be safely revealed and which must remain confidential~\cite{AlZubaidy2022}. This figure highlights the core challenge of responsible AI governance: maintaining individual privacy without compromising the transparency, explainability, or practical usefulness of AI systems.

Our framework addresses this challenge by structuring the feature space of individuals into two distinct groups: the \emph{open profile} and the \emph{private profile}. The open profile consists of features accessible to system observers, whereas the private profile contains features that must remain hidden. Additionally, we define a specific feature whose value must remain confidential to protect individuals who possess it, referred to as the \emph{sensitive feature}, and the corresponding confidential value as the \emph{sensitive literal}. The primary goal of our approach is to ensure that information in the open profile cannot be used to infer the sensitive literal ($s$), thereby preserving individual privacy while maintaining the usability of accessible features.

Having introduced this distinction, we proceed to formalize the notation and concepts underlying our frameworks.
\subsection{Notation}

We use the following notation throughout this paper.

\begin{itemize}
\item $V = \{v_1,\dots,v_n\}$ is a set of $n$ Boolean variables or \emph{features} that define individuals. 
\item An \emph{individual} $x = [x_1,\dots,x_n]$ is represented as a vector of Booleans associated with the $n$ features that define individuals.  
  The individual is also understood as a conjunction or a set of literals $v_i$ or $\neg v_i$.
  $x[i]$ refers to the literal $v_i$ or $\neg v_i$.
\item $X = 2^V$ is the space of all individuals.

\item $D$ is the domain space of the decision function and $d$ is an element of $D$:  $d \in D$.
\item $\Delta: X \to D$ is the \emph{decision function} of a machine learning model. 
\item  $V= V_{O} \cup V_{P}$ is a partition of $V$ into $V_{O}$, the set of \emph{open features}, and $V_{P}$, the set of \emph{private features}. We single out $s \in V_{P}$, the specific \emph{sensitive feature} that will violate individual privacy.  In particular, the observer shouldn't be able to determine when $x[s] = \nu$ for some specific value $\nu$.
\item $XP$ is any property of an individual $x$ defined as a subset (conjunction) of its literals. We use the following notation: $XP \subseteq x$.  We write $vars(XP)$ the features included in $XP$.
\item We use the notations $\oprof{S}$ and $\pprof{S}$ to represent the restriction of a set $S$ of literals to the open features and private features; for instance $\oprof{x}$ is the \emph{open profile} of $x$ and $\pprof{XP}$ is the private part of $XP$.
\item The property $XP \subseteq x$ is a \emph{valid explanation for $x$} if 
 \begin{displaymath}
            \forall x'.\ XP \subseteq x' \Rightarrow \Delta(x) = \Delta(x').
  \end{displaymath}
  When $x$ is implicit, we say that $XP$ is a \emph{valid explanation for~$\Delta$}.
  $XP$ is also called an \emph{abductive explanation}.
%
\item Explanation $XP$ is \emph{(subset-)minimal} if none of its strict subset is valid for the model.
\end{itemize}
\subsection{Access Levels and Framework Assumptions}
Analyzing privacy leakage requires a clear specification of the access levels of all system actors, which determines the actions permitted or prohibited for each~\cite{LI2015239}. In our framework, we consider two main roles: the \emph{Observer} and the \emph{Auditor}.

\paragraph{Observer Role.}  
The Observer has access to individuals' open profiles and their outcomes of the decision process. From a privacy perspective, this role is concerned with whether information in the open profile and outcomes could be exploited to infer private data. Observers may include system users verifying the safety of their own data or any actor with access to open profiles attempting to extract sensitive information.

\paragraph{Auditor Role.}  
The Auditor has access to decision process. The Auditor's goal is to ensure that privacy is always preserved in the decision process. In other words, to make sure that observers cannot use accessible information to infer any individual's sensitive literal.

Another key aspect of the framework is the assumption that all features are independent, preventing proxy effects. With these roles established, we now turn to defining the abductive explanations used in our framework for privacy leakage auditing.
\subsection{Abductive Explanations in Privacy Preserving Decision Systems}
Abductive explanation provide a formal method for reasoning about why a specific decision was made. By definition, a property $XP$ of a model is an abductive explanation if $XP$ is minimal and valid. Based on the configuration of our framework, we can distinguish several types of explanations, reflecting the relationship between the open and private profiles.

\begin{definition}
    An abductive explanation $XP$ is 
    \begin{itemize}
        \item an \textbf{open explanation} if all variables in $XP$ belong to the open profile.
        \item a \textbf{private explanation} if all variables in $XP$ belong to the private profile.
        \item is a \textbf{partial private explanation} if it includes variables from both profiles.
    \end{itemize}
\end{definition}

The categorization of different types of explanations opens the door to reflecting on how they may influence privacy leakage. One case, in particular, draws our attention: when an individual's decision admits an open explanation. This case is especially relevant because, for such an individual, there always exists at least one explanation that does not involve the sensitive literal. We refer to such a decision as a \emph{fully open decision}.

\begin{definition}[Fully Open Decision]
A decision $\Delta(x)$ is \emph{fully open} if it admits at least one open explanation.
\end{definition}

Fully open decisions are inherently privacy-safe, as their justification does not depend on the sensitive literal. Non–fully-open decisions, however, require a more nuanced auditing methodology, which we describe in the following sections.

\section{Privacy Leakage audit at the individual level}\label{sec::individual_level_audit}
Having established the foundational structure of our framework, we now turn to the analysis of \emph{privacy leakage at the individual level}. This section formalizes how privacy can be defined for a single individual within the decision process and presents the theoretical tools required to verify whether such privacy is preserved or compromised.

\subsection{Running Example}
To ensure clarity of the work, we introduce a minimal running example that will serve to illustrate our methodology of privacy leakage auditing. 
\begin{example}\label{example_description}
    Consider a decision process designed to determine whether tutors teaching undergraduate courses deserve supplements. The features of this decision process are:
    \begin{itemize}
        \item \textbf{Experience}: $E$ if more than 2 years, $\neg E$ otherwise;
        \item \textbf{Diploma}: $D$ if PhD student, $\neg D$ otherwise;
        \item \textbf{Medical condition}: $S$ if sick, $\neg S$ otherwise;
        \item \textbf{Extra hours}: $H$ if yes, $\neg H$ otherwise.
    \end{itemize}
\end{example}
In this example, we will consider the following information.
\begin{itemize}
    \item $V_O = \{E,D\}$; $V_P = \{S,H\}$.
    \item The sensitive literal is $S$, i.e., sick students do not want observers to know about their condition.
    \item $\Delta(x) = (x[D] \land (x[E]\lor x[H])) \lor x[S]$, i.e., a student will be supplemented if they are a PhD student who is either experienced or doing extra hours, or if they have a medical condition.
\end{itemize}

We consider multiple sick tutors who are all supplemented, and determine whether their condition is leaked by this decision:
\begin{itemize}
    \item $\textit{Toto} = E \land D \land S \land H$ is an experienced PhD student ($E \land D$).  As such, the outcome of the decision process was already known, so that no information leaks.
    \item $\textit{Tata} = E \land \neg D \land S \land H$ is not a PhD student.  An observer can therefore immediately conclude that the reason they were supplemented is their medical condition.  Sensitive information was leaked for \textit{Tata}.
    \item $\textit{Tintin} = \neg E \land D \land S \land H$ is a PhD student who does extra hours.  Their medical condition was not decisive in their outcome; in fact, $\textit{Tutu} = \neg E \land D \land \neg S \land H$---who is similar to \textit{Tintin} except that they are not sick---was also supplemented.
    \item $\textit{Tete} = \neg E \land D \land S \land \neg H$ is neither experienced nor doing extra hours; their medical condition was indeed decisive to their outcome.  However, from the observer's perspective, \textit{Tete} is undistinguishable from \textit{Tintin} or \textit{Tutu} (they are all non-experienced PhD student).  Thus, the observer cannot infer that \textit{Tete} has a medical condition.
\end{itemize}

In the rest of this section, we formalize the definition of private leakage 
and show that leakage is prevented when there exists a Potentially Applicable Explanation to the outcome
whose public part is applicable to the individual of interest.


\subsection{Formal Definition of Individual Privacy Leakage}
As a reminder, a central objective of our framework is to ensure that, for any individual $x$, information in the open profile cannot be used to infer the sensitive literal. Given that an observer has access to individuals' open profiles, this objective can be restated as follows: the framework must guarantee that the observer cannot exploit comparisons among individuals sharing the same open profile.

Indeed, if the observer can identify individuals with identical open profiles but different decisions, this discrepancy could reveal differences in their private profiles—precisely the type of inference our framework aims to prevent. Therefore, our ultimate goal is to preserve individual privacy by ensuring that individuals with similar open profiles receive similar outcomes. We formalize this requirement in the following definition.

\begin{figure}[H]
\centering
\includegraphics[scale=0.45]{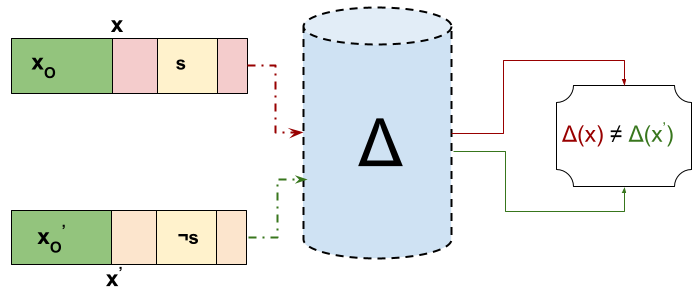}

\caption{Definition of privacy leakage}
\label{fig::privacy_leakage}
\end{figure}

\begin{definition}[Individual privacy preservation]
\label{defi::individual_privacy_leakage}
A decision process $\Delta$ does not leak sensitive information of an individual $x$ if
         \begin{displaymath}
             \exists x' \in X.\ \oprof{x} = \oprof{x'} \land x[s] \neq x'[s] \land \Delta(x)=\Delta(x')
         \end{displaymath}
\end{definition}
Definition \ref{defi::individual_privacy_leakage} states that a decision process does not leak for an individual $x$ if there exists an individual $x'$ with the same open profile as $x$ and differing on the sensitive literals that receive the same outcome as $x$. Our framework aims at detecting when this requirement is not satisfied as shown in figure~\ref{fig::privacy_leakage}.

The condition in Definition~\ref{defi::individual_privacy_leakage} can be equivalently formulated as follows: the decision process $\Delta$ leaks sensitive information about individual $x$
if this condition holds:
\begin{displaymath}
    \forall x' \in X. \oprof{x} = \oprof{x'} \land \Delta(x)=\Delta(x') \Rightarrow x[s]=x'[s],
\end{displaymath}
i.e., any individual with an open profile and a decision similar to $x$ must have the sensitive feature of $x$.

\subsection{Connection between Privacy Leakage and Abductive Explanation}

Let us recall that an abductive explanation is a minimal valid property of the model that ensures all individuals satisfying it receive the same outcome. Linking this concept to Definition~\ref{defi::individual_privacy_leakage}, we can observe that privacy leakage is avoided when two individuals $x$ and $x'$ sharing the same open profile satisfy the same valid property of the model. In cases where an individuals decision admit fully open explanation, there is no risk of privacy leakage, as there will always exist an explanation $XP$ that justifies their decisions without disclosing the \emph{sensitive} literal.

When this condition does not hold, the objective becomes identifying an explanation that can plausibly account for the decisions of both $x$ and $x'$. Since the observer has access only to open profiles, the open literals of such an explanation must be satisfied by both individuals. We formalize this idea through the notion of a \emph{Potentially Applicable Explanation (PAE)}:

\begin{definition}[Potentially Applicable Explanation (PAE)]\label{def::PAE}
A \emph{Potentially Application Explanation for $x$} is an explanation $XP$ for $\Delta$
such that $XP_O \subseteq \oprof{x}$ and $XP$ leads to the same decision as $x$.
\end{definition}

Definition~\ref{def::PAE} explicitly states that, within the framework of privacy leakage, an abductive explanation valid for the model can serve to justify an individual’s decision whenever the individual’s open profile satisfies the explanation. The next challenge is to demonstrate how this notion can be leveraged to systematically audit individual privacy leakage, which we address in the following section.

\subsection{Theoretical Foundations for Individual Leakage Auditing}

Building on the notion of $PAE$ this subsection discusses the main theorem that enables the formal auditing of privacy leakage at the individual level. Connecting Definitions~\ref{defi::individual_privacy_leakage} and~\ref{def::PAE}, the only condition to maintain an individual privacy is that the $PAE$ for its decision should not contain the sensitive literal. Otherwise, it will not be possible to use it to explain the decisions of two individuals that differs on the sensitive literal. Hence , ensuring privacy preservation of an individual is to find out a $PAE$ for its decision that does not involve the sensitive literal $s$. Thus we can state the following theorem.

\begin{theorem}\label{theo::no_leakage_individual}
    There is a $PAE$ for $x$ that does not include the sensitive literal iff  there is no leakage for $x$.
\end{theorem}
\paragraph{Proof:} 
 \paragraph{(\(\Rightarrow\))} 
 At first, let us prove that If there is a PAE for $x$ that does not include the sensitive literal, then there is no leakage for $x$

 Let us assume that $XP$ is a $PAE$ for $x$ and does not include the sensitive literal.

 $\Rightarrow$ Let us choose $x'$, an individual that differs from $x$ on exactly all private features/literals that are mentioned in $XP$ and that disagree with $x$. Because $XP$ is a $PAE$ for $x$ and an explanation for $x'$'s decision, $d=d'$. As a conclusion, 
 \begin{displaymath}
      \exists x' . \oprof{x} = \oprof{x'} \land d=d' \land x[s] \neq x'[s] 
 \end{displaymath}
Which is a definition of no leakage.

 \paragraph{(\(\Leftarrow\))} 
 At second , let us prove that If there is no leakage for $x$, then there exists a PAE for $x$ that does not include the sensitive literal.

Let us assume that there is \emph{no leakage} for $x$. This means 
\[
\exists x'.\; \oprof{x} = \oprof{x'} \land d = d' \land x[s] \neq x'[s].
\]
Let us take $XP$, an explanation of $x'$'s decision. Since $\oprof{x} = \oprof{x'}$, it follows that 
\[
XP_O \cap \oprof{x} = XP_O \cap \oprof{x'}.
\]
Moreover, because $XP \subseteq x'$ and $x'[s] \neq x[s]$, the explanation $XP$ excludes the sensitive literal $x[s]$. 

Therefore, we have 
\[
\exists XP \models \Delta(x') 
\;\;\text{such that}\;\; 
XP_O \cap \oprof{x} = XP_O \cap \oprof{x'} 
\land x[s] \notin XP.
\]

By definition, $XP$ is thus a PAE for $x$ that does not contain the sensitive literal $x[s]$.

We now connect the concept of PAE to the analysis of privacy leakage in case where an individual decision is \emph{fully open}. 

\begin{corollary}
    If an individual $x$ has a fully open decision, then there is always a PAE for x that does not include the sensitive literal.
\end{corollary}

\paragraph{Proof:} 
Let us assume that $x$ has a fully open decision. 
This means that $x$'s decision admits an explanation $XP$ such that 
\[
XP \cap \pprof{x} = \emptyset.
\] 
Thus $XP$ is a PAE for $x$. Plus, since it is fully open, it does not mention $p$. Thus, the condition of Theorem~\ref{theo::no_leakage_individual} is satisfied.

This section allows us to claim that an individual does not suffer from privacy leakage if and only if their decision admits a PAE that excludes the sensitive literal. As recall, a $PAE$ for $x$'s decision exists if there exists another individual $x'$ such that:
\begin{itemize}
     \item $x$ and $x'$ have the same open profile
    \item $x[s] \neq x'[s]$
    \item $PAE$ is an explanation of $x'$'s decision
\end{itemize}
Thus we can then propose a new concept called Leakage-Protected PAE (LPPAE), which ensures privacy preservation for individual decision.

\begin{definition}[Leakage-Protected PAE]\label{def::lppae}
A \emph{LPPAE} for an individual decision is a PAE that excludes the sensitive literal.
\end{definition}

Furthermore, by the preceding result, whenever an individual’s decision is fully open, such $LPPAE$ necessarily exists, ensuring the absence of privacy leakage. Let us return to the case of \emph{Tonton} =$E \land D\land S \land \neg H$, another individual from example~\ref{example_description} to check whether its sensitive information is leaked by the decision process.

\begin{example}\label{ex::toto_privacy_audit}
    Let us construct the individual \emph{Titi} with the same open profile as \emph{Tonton} but differing in all features of the private profile:\emph{Titi} = $E \land D \land \neg S \land H$ and also receives a positive outcome from the decision process $\Delta$, as it satisfies $XP = D \land H$, another valid property of the decision function. Consequently, $XP$ is a $PAE$ for \emph{Tonton}'s decision. Furthermore, $XP$ does not include the sensitive literal; thus, $XP$ qualifies as a $LPPAE$ for \emph{Tonton}'s decision. This condition suffices to conclude that \emph{Tonton}'s sensitive information is preserved by the decision process.
\end{example}

Example~\ref{ex::toto_privacy_audit} illustrates how privacy leakage can be audited at the individual level. In summary, auditing privacy leakage at the individual level is to check if its decision admit a $LPPAE$. Let us conclude this section by proposing an algorithm that could adress this challenge.
\begin{algorithm}[tb]
\caption{Search-LPPAE}
\label{alg:lppae}
\begin{algorithmic}[1]
\STATE
\textbf{Input}: $x$, an individual\\
\textbf{Input}: $\Delta$ The decision process\\
\textbf{Input}: $d$, individual decision
\STATE $XP\gets None$
\STATE $x'\gets SAT(d = d' \quad \land \quad \oprof{x} = \oprof{x'} \quad \land \quad x[s]\neq x'[s])$
\IF{$x' \neq$ None}
    \STATE Compute $XP$ such that $x'[s] \notin XP$ 
\ENDIF
\STATE \textbf{return} $XP$
\end{algorithmic}
\end{algorithm}

Algorithm~\ref{alg:lppae} defines a function that returns a $LPPAE$ for a given individual $x$, provided as input. To compute $XP$, the function first checks (line~2) whether there exists an individual $x'$ who shares the same open profile as $x$, receives the same decision, and differs on the sensitive literal. If such an individual exists, the function then computes (line~4) an abductive explanation for $x'$'s decision under the constraint that it excludes the sensitive literal $p$. If no such individual is found, the function returns \texttt{None}, indicating that the input individual $x$ suffers from privacy leakage.

Having established how privacy leakage can be audited at the individual level, we now extend this approach to ensure privacy preservation across the entire decision process, as discussed in the following section.

\section{Privacy Leakage Audit of a Decision Process}\label{sec::global_level_audit}

Building on our previous analysis of privacy leakage at the individual level using abductive explanations, we now extend our focus to privacy risks at the level of the entire decision process. 

\begin{definition}[Model leakage]\label{def::model_leakage}
    The model \emph{suffers from leakage} if there is an individual for which the model leaks the sensitive literal.
\end{definition}
Definition~\ref{def::model_leakage} states that the model is leaking if it is only possible to find out one individual that suffers from privacy leakage. In other words, to prove model leakage, man should demonstrate a specific case of an individual suffering from privacy leakage. Thus the central question to audit a decision process is how to make sure that none individual is suffering from privacy leakage?

This section is organized in four parts: First, we discuss the complexity checking privacy leakage.  Second, we formalize the notion of privacy leakage within the model.  Third, we propose an algorithm that leverages this formalization to audit the entire decision process and finally we evaluate the algorithmic complexity of the proposed methodology.

\subsection{Computational Complexity}

We now discuss the complexity of deciding if a decision process is leaking.

\begin{definition}
    A \emph{Leaking Instance} is defined as a tuple ${\mathcal P} = \langle V, V_O, V_P, s, \nu, \Delta\rangle$
    whose elements are defined as above.
    The \emph{Leaking Problem} is to decide if $\Delta$ is leaking information for some individual.
\end{definition}

\newcommand{\sigmaptwo}{$\Sigma^{\textnormal{P}}_2$}

\begin{lemma}\label{lemma::in}
    The leaking problem is in \sigmaptwo.
\end{lemma}

This is trivial since the problem is defined as a $\exists\forall$ problem over a finite set of Boolean variables.

\begin{lemma}\label{lemma::hard}
    The leaking problem is \sigmaptwo-hard.
\end{lemma}

Consider the $\exists\forall$SAT problem $\exists Y.\ \forall Z.\ \varphi(Y,Z)$, which is known to be \sigmaptwo-complete.
We define the following leakage problem ${\cal P}$:

\begin{itemize}
\item The set of features is $V = Y \cup Z \cup \{s\}$ where
  $Y$ is the set of open features, $Z \cup \{s\}$ is the set of private features 
  and $s$ is the sensitive feature.
\item The decision procedure is $\Delta(x) = x[s] \lor \neg \varphi(x)$.
\item The sensitive literal is $s = True$.
\end{itemize}

If $\Delta$ is leaking,
then there exists $x$ such that for all $x'$
that verify $x_O = x'_O$, $x[s] = True$, and $x'[s] = False$,
then $\Delta(x) \neq \Delta(x')$.
Because of the definition of $\Delta$ (and since $x'[s] \neq True$),
$\varphi(x')$ is false.
Furthermore, since $V_O = Y$, this implies that we found an assignment $y$ of $Y$
such that for all assignment $z$ of $Z$, $\varphi(y,z)$ evaluates to True,
i.e., a solution to the $\exists\forall$SAT problem.

Conversely, it is easy to see that any solution to the $\exists\forall$SAT
yields a class of individuals $x$ (such that $x_O = Y$ and $x[s] = True$)
to the leaking problem $P$.

\begin{corollary}\label{coro::sigmaptwo}
    The leaking problem is \sigmaptwo-complete.
\end{corollary}

Corollary~\ref{coro::sigmaptwo} is a direct consequence of Lemmas~\ref{lemma::in} and~\ref{lemma::hard}.

\subsection{From Individual to Model-Level Privacy Preservation}

A naive approach would be to examine every individual decision to check for potential privacy leakage. While correct in principle, such an approach would be prohibitively expensive in terms of time and computational resources. We therefore seek methods that reduce this complexity. To this end, we revisit the analysis of the $LPPAE$ criterion under the following scenario.

Consider two individuals, $x$ and $x'$, who share the same open profile and receive the same decision but differ in their private profile. Suppose that $x$ does not suffer from privacy leakage; that is, $x$ admits an abductive explanation $XP$ satisfying the $LPPAE$ criterion that justifies and protects its decision. Since $x$ and $x'$ share the same open profile, $XP$ will also serve as an $LPPAE$ for the decision of $x'$. This observation leads to the following theorem:

\begin{theorem}\label{theo::lppae_x_and_x'}
If two individuals $x$ and $x'$ share the same public profile, have the same decision, and $XP$ is a $LPPAE$ for $x$, then $XP$ is also a $LPPAE$ for $x'$.
\end{theorem}
\paragraph{Proof}
let $x$ and $x'$ be two individuals with the same decision such that $\oprof{x} = \oprof{x'}$ and $XP$
be $LPPAE$ for $x$. This means that
\begin{displaymath}
    \oprof{x} \cap XP = \oprof{x'} \cap XP = XP_O
\end{displaymath}
 Thus, $XP$ is a $PAE$ for $x'$. Since $XP$ excludes the sensitive literal, $XP$ is also a $LPPAE$ for $x'$.

Theorem~\ref{theo::lppae_x_and_x'} establishes that for individuals sharing the same public profile and the same outcome, a single $LPPAE$ is sufficient to determine whether they experience privacy leakage. This result implies that, by partitioning the set of individuals according to their public profiles, the task of detecting model-level leakage can be reduced to verifying the existence of an $LPPAE$ for each group. Building on this observation, we introduce the following definition.

\begin{definition}[LPPAE-cluster]\label{def::lppae_cluster}
A $LPPAE$-cluster, denoted $X^x_{V_P}$, is the set of all individuals sharing the same public profile and same decision as $x$.
\end{definition}
Definition~\ref{def::lppae_cluster} allows us to partition the set of individuals into clusters covered by the same $LPPAE$, a criterion of leakage protection for individuals sharing the same open profile and decision. This represents a meaningful step in reducing the complexity of the naive assessment method described earlier. 

However, due to the persuasive property of a $LPPAE$, we can claim that to be covered by the same $LPPAE$ as $x$, an individual does not need to share the same public profile; it suffices to satisfy the open literals included in the $LPPAE$ of $x$. This refinement enables a more efficient partitioning of individuals: a single $LPPAE$ can cover a larger set of individuals, further reducing the complexity of the naive method. We therefore introduce the following definition.

\begin{definition}[LPPAE-equivalence class]\label{def::lppae_equivalence_class}
The $LPPAE$-equivalence class of $XP$, denoted $X_{(XP_O)}$, is the set of all individuals that satisfy the literals contained in $XP_O$ and having the same decision.
\end{definition}

In summary, auditing a decision process consists of verifying whether an $LPPAE$ exists for each $LPPAE$-equivalence class. The model is said to suffer from privacy leakage if it is not possible to assign an $LPPAE$ to every individual. Based on Definition~\ref{def::lppae_equivalence_class}, the next section presents an algorithm that operationalizes this methodology.

\subsection{Auditing Privacy Leakage in the Decision Process}
At this stage, we have shown that auditing can be made more efficient by identifying $LPPAE$-equivalence classes that guarantee privacy preservation for a larger set of individuals. In the following, we present algorithms that exploit these equivalence classes to accelerate the verification of privacy leakage in the decision process.

\begin{algorithm}[tb]
\caption{Leakage Detection with LPPAE}
\label{alg::leakage_detection}
\begin{algorithmic}[1]
\STATE
\textbf{Input}: $\Phi(c)$ SAT formula that return the individuals that satisfy constraint $c$ \\
\textbf{Input}: $K$ Constraints\\
\textbf{Input}: $\Delta$ the decision process
\STATE Update $K$ AND $p$
\WHILE{$\Phi(K) = SAT$}
    \STATE Choose $x \in \Phi(K)$.
    \STATE$d\gets \Delta(x)$
    \STATE $XP \gets$ Search-LPPAE($x$, $\Delta$, d)
    \IF{$XP = None$}
        \STATE \textbf{return} leakage detected.
    \ENDIF
    \STATE Update $K$ AND NOT $XP_O$
\ENDWHILE
\STATE \textbf{return} no leakage.
\end{algorithmic}
\end{algorithm}

Algorithm~\ref{alg::leakage_detection} outlines the steps to assess privacy leakage in a decision process. The inputs to the algorithm are as follows: $\Phi(c)$, a SAT formula that returns all individuals satisfying the constraint $c$; $K$, a set of constraints to be verified during the audit; and $\Delta$, the decision process itself, which is subject to auditing.

At line~1, $K$ is initialized with the sensitive literal $p$, so that the audit only considers individuals who could be subject to privacy leakage; individuals not satisfying this literal are by default exempt. From line~2, the algorithm iteratively checks whether there exists an individual satisfying the current constraints in $K$ who has not yet been verified. If such an individual exists, the Algorithm~\ref{alg:lppae} is called to determine whether the individual's decision admits an $LPPAE$. If the search is negative, the algorithm stops (line~7), as it has identified an individual suffering from privacy leakage, which suffices to conclude that the model leaks. 

If the search is positive, the constraint set $K$ is updated with the open literals of the found $LPPAE$ (line~9) to ensure that the next candidate does not belong to an $LPPAE$-equivalence class already covered. The algorithm continues until either all individuals are covered and the decision process is free of privacy leakage, or it stops early upon detecting an individual with privacy leakage.

\subsection{Algorithmic Complexity of Model Leakage Auditing}
To analyze the computational complexity of Algorithm~\ref{alg::leakage_detection}, we consider the following aspects. Recall that $D$ denotes the domain space of the decision process, and $V_O$ represents the set of variables included in the open profile of individuals. To estimate the worst-case complexity, we assume the scenario in which an $LPPAE$ exists for an individual $x$ and involves all variables from the open profile. 

Before computing an $LPPAE$, the algorithm must perform at least two mandatory operations: (i) select an individual $x$ that has not yet been audited, and (ii) generate its corresponding $x'$, which shares the same open profile and decision as $x$ but differs on the sensitive literal $p$. Therefore, the overall computational complexity of the algorithm is given by $|D| \times 2^{(2 + |V_O|)}$.

This exponential complexity highlights that the audit process may become computationally demanding as the number of observable features grows, emphasizing the need for optimization strategies or approximation methods in large-scale decision systems.

\section{Experimental Study}\label{sec::experimental}
To empirically evaluate our proposed framework, we conduct a series of experiments using the \emph{German Credit} dataset. In this setting, the decision process is modeled as a binary classifier that determines whether an individual is assessed as having a good or bad credit score, simulating a loan approval scenario. The objective of this study is to evaluate which models exhibit privacy leakage, as well as to measure the time required to assess such leakage across different classifiers.

This section is organized into three parts. At first we describe the dataset and details the separation between open and private profiles, as well as the identification of the sensitive literal. At second, we present the classification models used in the experiments namely, a neural network (\textbf{M1}), a logistic regression model (\textbf{M2}), and an SGD classifier (\textbf{M3}). Finally, we reports and analyzes the results, highlighting how each model behaves with respect to privacy leakage and auditing efficiency.

\subsection{Description of the Dataset}
The \emph{German Credit} dataset (GCD) consists of the following features: Age ($A$), Gender ($G$), Job ($J$), Housing ($H$), Savings ($S$), Account Balance ($B$), Credit Amount ($C$), Duration of Last Credit ($D$), Credit Purpose ($P$), and Marital Status ($M$). 

In this study, we consider the following configuration for the \emph{open profile} and the \emph{private profile}:
\begin{itemize}
    \item $V_O = \{A, G, J, H, B\}$
    \item $V_P = \{S, C, D, P, M\}$
\end{itemize}

The \emph{Credit Purpose} ($P$) attribute is designated as the sensitive variable, taking the value $P = 1$ if the purpose of the credit is a car, and $P = 0$ (or $\neg P$) otherwise. Since the dataset includes specific instances where the credit purpose is a car, we use $\neg P$ as the sensitive literal in our analysis.

\subsection{Decision Processes Used }
As discussed earlier, our experimental setup includes three models with distinct learning characteristics, as summarized in Table~\ref{tab::models}.
\begin{table}[h!]
\centering
\begin{tabular}{|c|c|c|c|c|}
\hline
Models & Accuracy & Precision & Recall & F1-score \\ \hline
    \textbf{M1}   &   0.74       &  0.78      &   0.87       &    0.82      \\ \hline
    \textbf{M2}   &     0.71     &  0.74     &       0.91   &       0.82   \\ \hline
    \textbf{M3} &        0.72  &     0.75     &       0.89   &       0.82   \\ \hline
\end{tabular}
\caption{Characteristics of the models on the GCD}
\label{tab::models}
\end{table}
These models are converted into SMT formulas, which then serve as input to Algorithm~\ref{alg::leakage_detection} described above. The results of this procedure are presented and analyzed in the next section.
\subsection{Results}

As reported in Table~\ref{tab::results_leakage}, models \textbf{M1} and \textbf{M2} show no signs of privacy leakage, in contrast to model \textbf{M3}, which is detected as leaking under our framework. The audit process is highly efficient for all models, completing in under 10 seconds, with \textbf{M3} despite its leakage being the fastest to evaluate (0.09s).

\begin{table}[h!]
\centering
\begin{tabular}{|c|c|c|}
\hline
Models & Leaking & Runtime (s) \\ \hline
     \textbf{M1}       &      No   &   9.71        \\ \hline
    \textbf{M2}        &       No  &      0.11    \\ \hline
     \textbf{M3}       &       Yes   &     0.09     \\ \hline
\end{tabular}
\caption{Results of the experiments}
\label{tab::results_leakage}
\end{table}

\section{Discussion}
This work introduces a novel methodology to assess privacy leakage in AI-based decision processes by leveraging abductive explanations. Within our framework, an individual’s sensitive information is considered preserved if there exists another individual who does not exhibit the sensitive literal but receives the same outcome as $x$; in this case, we say that $x'$ is \emph{shielding} $x$. Privacy is maintained across the entire decision process if every individual possessing the sensitive literal is assigned a shield. 

In this section, we discuss three key aspects of the framework. First, we analyze its efficiency in light of the computational complexity of the privacy leakage problem. Second, we examine the relationship between the accuracy of the decision process and privacy preservation. Finally, we identify potential avenues to improve and extend the framework for broader applicability.

\subsection{Computational Complexity of Privacy Leakage Auditing}
The computational complexity of the privacy leakage problem has been proved to be  \sigmaptwo-complete.
This result highlights both the theoretical depth and computational challenge of auditing privacy leakage. Being $\Sigma_{2}^{P}$-complete, the problem involves reasoning over alternating quantifiers and is generally intractable. Nevertheless, this complexity motivates principled frameworks like ours: abductive inference allows the identification of leakage patterns via minimal explanations, providing a practical and interpretable approach to an otherwise intractable task.
\subsection{Impact of the Sensitive Literal on Privacy Leakage}

From Table~\ref{tab::results_leakage}, we observe that model leakage is detected more quickly than non-leaking cases, which can be explained by Algorithm~\ref{alg::leakage_detection}: when leakage occurs, the algorithm requires fewer iterations. However, there is no direct correlation between model accuracy and the presence of privacy leakage. For instance, in our experiments, \textbf{M1} and \textbf{M2} do not leak, yet their auditing times differ significantly. Additionally, \textbf{M2} has lower accuracy than \textbf{M3}, but only \textbf{M3} exhibits leakage.  

These observations suggest that privacy leakage is closely tied to the sensitivity of the decision process to the protected literal: the more a sensitive feature influences the outcome, the more likely the model is to leak. This aligns with the work of Darwiche and Hirth~\cite{Adnan2020Reasons}, who define an individual decision as biased with respect to a feature if all explanations justifying that decision include the feature. Within our framework, we can formally show that if an individual suffers from privacy leakage, all explanations related to its decision include the protected literal.
\begin{corollary}
If an individual $x$ suffers from privacy leakage, then all explanations of $x$'s decision include the sensitive literal $s$.
\end{corollary}

\paragraph{Proof}
If $x$ suffers from privacy leakage, there exists no $LPPAE$ for its decision. This implies that no individual with the same open profile but a different sensitive literal can be found that receives the same decision. Consequently, every explanation justifying $x$'s decision must include the sensitive literal $s$.

\subsection{Limitations and Future Work}

While our framework provides strong formal guarantees based on abductive reasoning and produces human-understandable explanations, a key requirement for auditing tools, it is computationally demanding. Additionally, the framework relies on the strong assumption that all features are independent, which simplifies the analysis but may not hold in real-world scenarios. This assumption was adopted to avoid the complexity introduced by proxy features, which has been addressed in previous work on abductive explanations~\cite{sonna2025explainingproxydiscriminationunfairness}. Despite these limitations, the framework ensures that privacy leakage can be formally identified and interpreted, offering a principled foundation for privacy-aware auditing.

Our experimental evaluation, conducted on the GCD, illustrates the framework in a controlled setting; however, this dataset is relatively simple and does not capture the full complexity of real-world decision processes. Future work will focus on extending the framework to more complex datasets and relaxing the independence assumption to handle proxies features. Another promising direction is to develop methods to sanitize decision processes, mitigating privacy leakage while preserving model accuracy, thereby providing actionable tools for privacy-preserving AI deployment.
\section{Conclusion}
We have presented a novel framework for auditing privacy leakage in AI-based decision processes leveraging abductive explanations. By formally reasoning about minimal evidence, we identify a property of the model ($PAE$) that could allow the use of an individual without sensitive information to protect the one which actually possess an sensitive information. In addition, our approach provides rigorous privacy guarantees while generating interpretable outputs for human auditors. The framework successfully detects privacy leakage at both individual and system levels, as demonstrated in our experimental study on the GCD. 

Despite computational challenges and simplifying assumptions the framework establishes a principled foundation for transparent privacy-aware auditing. Future work will extend the methodology to more complex datasets, handle correlated features, and explore strategies to sanitize decision processes to mitigate leakage without compromising model accuracy. Overall, this work highlights the potential of abductive reasoning to reconcile transparency, interpretability, and privacy in AI systems, contributing to more accountable and trustworthy decision-making.
\bibliography{aaai2026}

\end{document}